%% file: AnonymousSubmission2027.tex
\documentclass[letterpaper]{article} % DO NOT CHANGE THIS
\usepackage{aaai2027}
\usepackage[hyphens]{url}  % DO NOT CHANGE THIS
\usepackage{graphicx} % DO NOT CHANGE THIS
\usepackage{natbib}  % DO NOT CHANGE THIS AND DO NOT ADD ANY OPTIONS TO IT
\usepackage{caption} % DO NOT CHANGE THIS AND DO NOT ADD ANY OPTIONS TO IT
\usepackage{algorithm}
\usepackage{algorithmic}
\usepackage{amsmath}
\usepackage{bm}
\usepackage{amssymb}
\usepackage{multirow}
\usepackage[table]{xcolor}
\usepackage{booktabs}

\definecolor{baselinecolor}{gray}{.9}
\newcommand{\cc}[1]{\cellcolor{baselinecolor}{#1}}

\usepackage{newfloat}
\usepackage{listings}
\DeclareCaptionStyle{ruled}{labelfont=normalfont,labelsep=colon,strut=off} % DO NOT CHANGE THIS
\floatstyle{ruled}
\newfloat{listing}{tb}{lst}{}
\floatname{listing}{Listing}

\usepackage{booktabs}

\title{Coupled Continuous–Discrete Generation for Scene Text Image Super-Resolution}

\author{
    Axi Niu\textsuperscript{\rm 1},
    Kang Zhang\textsuperscript{\rm 2},
    Qingsen Yan\textsuperscript{\rm 1},\\
    Hao Jin\textsuperscript{\rm 1},
    Jinqiu Sun\textsuperscript{\rm 1},
    Yangning Zhang\textsuperscript{\rm 1}
}

\affiliations{
    \textsuperscript{\rm 1}
    Northwestern Polytechnical University, Xi'an, China\\
    \textsuperscript{\rm 2}
    Korea Advanced Institute of Science and Technology (KAIST),
    Daejeon, Republic of Korea\\
}

\begin{document}

\maketitle

\input{sec/0_abstract}

\input{sec/1_intro}
\input{sec/2_related_work}

\input{sec/3_method}

\input{sec/experiment_new}
\input{sec/5_conclusion}


% \bibliography{aaai2027}

\bibliography{main}

\input{sec/6_appendix}

\end{document}

%% file: sec/0_abstract.tex
\begin{abstract}

Scene text image super-resolution (STISR) aims to recover visually plausible appearance while preserving character semantics from degraded inputs. Existing STISR systems often rely on externally generated priors or separate image and text models, resulting in error propagation and costly multi-stage inference. We present DualTSR, a unified framework that formulates STISR as coupled continuous--discrete generation. Conditional flow matching restores continuous image latents, while absorbing-state discrete diffusion reconstructs text tokens. Both processes share a multimodal transformer backbone, allowing the evolving image and text states to interact throughout generation without an external OCR prior at inference. On CTR-TSR, DualTSR achieves the best FID, LPIPS, ACC, and NED among the compared methods at both $\times2$ and $\times4$. On an aligned RealCE subset, it obtains the best FID, ACC, and NED with competitive LPIPS. Compared with DiffTSR at $\times4$, DualTSR improves ACC by 12.78 percentage points while reducing the parameter count from 1.23B to 203M and end-to-end latency from 13.3s to 132ms. These results establish DualTSR as an accurate and efficient method for STISR.

\end{abstract}

%% file: sec/1_intro.tex
\section{Introduction}
\label{sec:intro}

Scene Text Image Super-Resolution (STISR) aims to reconstruct high-quality scene text images from low-resolution inputs. The task is especially challenging because character structures are sensitive to degradation: small distortions, missing strokes, or blending with background clutter can alter the underlying semantics. Unlike general image super-resolution, which primarily enhances textures, STISR must simultaneously guarantee \emph{textual fidelity} and \emph{stylistic realism}. Any structural error directly causes recognition failure, while inconsistencies in font, color, or orientation reduce visual plausibility. These difficulties are further amplified in languages with large character sets such as Chinese.

To reduce the inherent ambiguity of STISR, recent work~\cite{chen2021scene, ma2023text, ma2023benchmark, wang2020scene} incorporates textual priors extracted from a pre-trained OCR model. Knowing the expected text sequence significantly constrains the solution space and often improves reconstruction quality. This OCR-guided paradigm has therefore become dominant, and several recent methods~\cite{noguchi2024scene, zhang2024diffusion} explicitly condition the SR process on predicted text tokens to enforce semantic correctness. However, its reliability is fundamentally limited by the accuracy of the external OCR: erroneous predictions propagate into the SR network, leading to hallucinated strokes, incorrect glyphs, and overall degraded results. To mitigate this dependency, subsequent works introduced stronger low-level priors. MARCONet~\cite{li2023learning} learns a discrete codebook of character structures to provide robust structural guidance, while GlyphSR~\cite{wei2025glyphsr} goes further by generating explicit glyph masks using SAM to supervise fine-grained stroke reconstruction. Although these structural priors alleviate the brittleness of OCR guidance, they require increasingly complex and multi-stage pipelines for prior extraction. This growing reliance on external modules highlights a core limitation of existing approaches and motivates a unified framework that can learn both semantic and structural cues internally without handcrafted or externally produced priors.

To this end, diffusion-based generative models~\cite{songdenoising, lipman2023flow} have recently been adopted for STISR, particularly for complex scripts~\cite{noguchi2024scene, zhang2024diffusion}. DiffTSR~\cite{zhang2024diffusion} reports strong results by modeling text and image with separate modules connected through a multi-modality fusion block. While effective, this multi-stage architecture is cumbersome: it increases system size, complicates training, and limits the depth of cross-modal interaction because textual and visual information are only exchanged at specific fusion points.

To address these limitations, we propose \texttt{DualTSR}, a unified STISR framework that formulates restoration as coupled continuous-discrete generation. Conditional flow matching generates the high-resolution image latent, while discrete diffusion reconstructs the corresponding character sequence. Both processes are parameterized by a shared multimodal transformer, allowing image and text representations to interact throughout the network. Instead of receiving a text sequence from an external OCR model, DualTSR predicts the text internally from the low-resolution observation and the evolving image state. During training, synchronized corruption presents partially corrupted image and text states at a shared timestep. During inference, the shared backbone updates both states jointly, allowing the current text hypothesis to guide glyph restoration and the evolving image to refine text prediction. \texttt{DualTSR} therefore replaces separate image and text generative backbones with a compact joint formulation.

Our contributions are as follows:
\begin{itemize}
    \item We introduce \texttt{DualTSR}, a unified STISR framework that jointly models continuous image restoration and discrete text prediction within a shared multimodal transformer, without requiring an external OCR prior at inference.
    \item We formulate a coupled training and inference scheme that combines conditional flow matching for image generation with absorbing-state discrete diffusion for text prediction. Synchronized corruption and joint updates enable the evolving image and text states to condition each other throughout generation.

    \item \texttt{DualTSR} achieves the best FID, LPIPS, ACC, and NED among the compared methods on CTR-TSR at both $\times2$ and $\times4$, and the best FID, ACC, and NED with competitive LPIPS on the aligned RealCE subset. Compared with DiffTSR at $\times4$, it improves ACC by 12.78 percentage points while using $6.1\times$ fewer parameters, $4.5\times$ less peak memory, and running approximately $101\times$ faster.

\end{itemize}

%% file: sec/2_related_work.tex
\begin{figure*}[t]
\begin{center}
 \includegraphics[width= \linewidth]{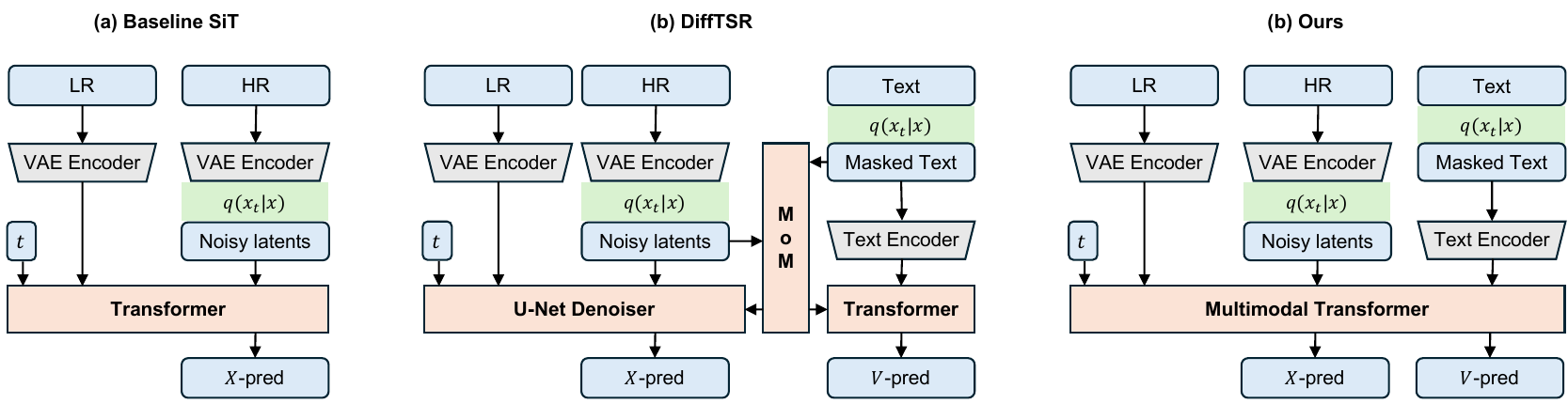}
\end{center}
\caption{\textbf{Comparison of representative STISR architectures.} DualTSR unifies image generation and text prediction within a single model.
(a) Baseline: Directly models the conditional distribution of high-resolution images from low-resolution inputs, without a text-based prior.
(b) DiffTSR~\cite{zhang2024diffusion}: Independently models the image and text priors, subsequently fusing their representations using a Multi-modality Module (MoM).
(c) DualTSR: A unified multimodal transformer jointly optimizes image generation and text prediction through two coupled diffusion processes.}
\label{fig:framework}
\end{figure*}

\section{Related Work}
\subsection{Text Image Super-Resolution}
Research in Text Image Super-Resolution (TISR) addresses the unique challenges of simultaneously enhancing image details and improving text legibility. While general-purpose blind super-resolution methods such as BSRGAN~\cite{zhang2021designing} and Real-ESRGAN~\cite{wang2021real} restore natural images well, they often fail to preserve the fine-grained structures and appearance cues required for text. This has motivated specialized STISR methods that introduce text-specific priors or auxiliary supervision.

Early and mid-stage STISR work explored complementary cues for restoration. Text Gestalt~\citep{chen2022text} emphasizes stroke-aware reconstruction, TATT~\citep{ma2022text} introduces text attention to improve robustness to spatial deformation, and C3-STISR~\citep{zhao2022c3} combines multiple clues to recover both semantics and structure. A large line of work then adopts OCR-guided priors, where a pre-trained recognizer predicts text that is fed into the SR network to enforce semantic correctness~\citep{wang2020scene, ma2023text, noguchi2024scene, zhang2024diffusion}. These methods can substantially improve recognition-oriented metrics, but their performance remains tied to the quality of the external OCR prior.

More recent methods instead emphasize structural priors. MARCONet~\citep{li2023learning} learns a character-structure codebook, MARCONet++~\citep{li2025enhanced} further strengthens this line with improved character priors, and GlyphSR~\citep{wei2025glyphsr} uses explicit glyph masks generated from an auxiliary pipeline. These approaches reduce dependence on raw OCR predictions, but they also introduce additional stages, modules, or prior-generation procedures. DualTSR differs from both families: rather than injecting a separately predicted text prior or a handcrafted structural prior, it learns image restoration and text prediction jointly within one multimodal backbone, so semantic cues are produced inside the model instead of being imported from a separate pipeline.

\subsection{Diffusion Models in Super-Resolution}
Diffusion probabilistic models have demonstrated state-of-the-art performance in image synthesis and restoration, owing to their powerful ability to model complex data distributions~\cite{ho2020denoising, lipman2023flow}. This success has naturally led to their application in both natural image super-resolution~\cite{niu2024acdmsr,shang2024resdiff,moser2024diffusion} and TISR ~\cite{zhao2024pean,zhou2024recognition,liu2025textdiff}. Furthermore, research has shown that diffusion models are not only suitable for continuous data like images but can also effectively model discrete data distributions, such as text sequences~\cite{austin2021structured,meng2022concrete,lou2023discrete,sahoo2024simple}.

In the context of TISR, diffusion models have been used in several ways to mitigate unreliable priors and ambiguous degradations. PEAN~\citep{zhao2024pean} refines text-aware features with diffusion-based priors, TextDiff~\citep{liu2023textdiff} uses mask-guided residual diffusion for restoration, and recognition-guided or text-conditional diffusion models~\citep{zhou2024recognition, noguchi2024scene} strengthen the coupling between restoration and text prediction. DiffTSR~\citep{zhang2024diffusion} pushes this direction further by separating image generation and text generation into dedicated diffusion branches connected by a multi-modality module, which is particularly effective for Chinese scene text. DualTSR keeps the central diffusion-based idea of joint image-text modeling, but replaces the modular design with a single multimodal transformer trained by two coupled objectives: flow matching for the image branch and discrete diffusion for the text branch. The key distinction is therefore architectural: we trade explicit cross-model fusion for shared representations and layer-wise multimodal interaction.

%% file: sec/3_method.tex
\section{Method}
We introduce \textbf{DualTSR}, an end-to-end multimodal diffusion framework for scene text image super-resolution. The central design choice is to model high-resolution image synthesis and text prediction inside a single shared backbone rather than through separately trained priors. Given a low-resolution input image $\mathbf{x}^{\text{lr}}$, our objective is to learn the conditional joint distribution $p(\mathbf{x}^{\text{hr}},\mathbf{x}^{\text{txt}}|\mathbf{x}^{\text{lr}})$, where $\mathbf{x}^{\text{hr}}$ denotes the restored high-resolution image and $\mathbf{x}^{\text{txt}}$ denotes the text sequence contained in that image. As illustrated in Figure~\ref{fig:framework}, DualTSR combines three components: (1) a conditional flow matching process for high-resolution image synthesis $p(\mathbf{x}^{\text{hr}}|\mathbf{x}^{\text{lr}},\mathbf{x}^{\text{txt}})$; (2) a discrete diffusion process for text prediction $p(\mathbf{x}^{\text{txt}}|\mathbf{x}^{\text{lr}},\mathbf{x}^{\text{hr}})$; and (3) a multimodal transformer that shares representations across both processes to learn the joint distribution end to end. This allows the evolving image and text states to inform one another throughout training and inference.

\subsection{Preliminaries}

Our model builds upon two key generative modeling paradigms: conditional flow matching for continuous data like images, and discrete diffusion for textual data.

\subsubsection{Conditional Flow Matching for Image Generation}
Flow matching~\citep{lipman2023flow, tong2024improving} is a recent generative modeling paradigm that formulates sample generation as solving an ordinary differential equation (ODE) defined by a time-dependent velocity field.
At inference, the goal is to generate a high-resolution image $\mathbf{x}_0$ by solving an ODE over a time interval $t \in [0, 1]$. The process starts with a random noise sample $\mathbf{x}_1$ drawn from a standard normal distribution, $p_1(\mathbf{x}_1) = \mathcal{N}(0, I)$. The ODE solver then integrates a learned conditional velocity field, $\mathbf{v}_\theta(\mathbf{x}_t, t, \mathbf{c})$, which guides the transformation from noise to an image. This velocity field is conditioned on external information $\mathbf{c}$ (in our case, the low-resolution image and optionally text labels) and is parameterized by the image head of our multimodal network.

During training, the network $\mathbf{v}_\theta$ is optimized to predict the velocity of a simple linear trajectory between a noise sample $\mathbf{x}_1$ and a target data sample $\mathbf{x}_0$. The path is defined as
\begin{equation}
\mathbf{x}_t = (1-t)\mathbf{x}_0 + t\mathbf{x}_1,
\label{eq:img_forward}
\end{equation}
and its corresponding constant velocity is $\bm{u}_t = \mathbf{x}_1 - \mathbf{x}_0$. The CFM objective minimizes the discrepancy between the predicted velocity and this ground-truth velocity:
\begin{equation}
\mathcal{L}_{\text{CFM}} = \mathbb{E}_{t, q(\mathbf{x}_1), q(\mathbf{x}_0, \mathbf{c})} \left\| \mathbf{v}_\theta(\mathbf{x}_t, t, \mathbf{c}) - \bm{u}_t \right\|^2
\label{eq:cfm_loss}
\end{equation}
where $t$ is sampled uniformly from $[0, 1]$, $q(\mathbf{x}_1)= \mathcal{N}(0, I)$ is the standard normal distribution, and $q(\mathbf{x}_0, \mathbf{c})$ samples from the training dataset.

\subsubsection{Discrete Diffusion for Text Generation}
\label{sec:discrete_diffusion}
For text generation, we require a diffusion process tailored for discrete data, where each token $\mathbf{x}$ belongs to a finite vocabulary $\mathcal{X} = \{1, \dots, N\}$. While some methods apply diffusion in a continuous latent space~\citep{li2022diffusion,chen2022analog,dieleman2022continuous,lovelace2024latent,gulrajani2024likelihood}, this can introduce mapping errors. We instead adopt a diffusion process that operates directly in the discrete token space, a paradigm that has shown strong empirical results \citep{austin2021structured,meng2022concrete,lou2023discrete,sahoo2024simple}.

Our approach is based on a continuous-time Markov chain (CTMC) that defines a forward corruption process. Specifically, we use an absorbing-state diffusion process~\citep{lou2023discrete, sahoo2024simple, shi2024maskdiff}, where tokens in the original text sequence $\mathbf{x}$ are progressively replaced by a special mask token $\mathbf{m}$. The marginal distribution of the corrupted text $\mathbf{x}_t$ at time $t$ is a categorical distribution conditioned on the original text $\mathbf{x}$:
\begin{equation}
    \label{eq:mdlm_xt}
    q(\mathbf{x}_t | \mathbf{x}) = \text{Cat}[\mathbf{x}_t | \alpha_t \mathbf{x} + (1 - \alpha_t) \mathbf{m}],
\end{equation}
where $\text{Cat}(\cdot|\boldsymbol{\pi})$ is the categorical distribution with probabilities $\boldsymbol{\pi}$, and $\alpha_t$ is a noise schedule.

The goal of the reverse process is to learn a denoising distribution $p_\theta(\mathbf{x}\mid \mathbf{x}_t, t)$ that predicts the original clean text $\mathbf{x}$ from its corrupted version $\mathbf{x}_t$. Following recent work \citep{sahoo2024simple, shi2024maskdiff}, we train the model by directly predicting the denoised variate. This leads to a simplified negative variational lower bound (NELBO) objective under the continuous-time limit:
\begin{equation}
\label{eq:mdlm_elbo}
    L_{\text{NELBO}} = \mathbb{E}_{q(\mathbf{x}_t|\mathbf{x})}
    \left[
    \int_{0}^1 \frac{-\alpha_t'}{1-\alpha_t}\log p_\theta(\mathbf{x}\mid \mathbf{x}_t, t)\, dt
    \right].
\end{equation}
In practice, this integral is approximated using Monte-Carlo sampling. For the noise schedule, we follow \citet{sahoo2024simple} and use a simple log-linear schedule where $\alpha_t = 1-t$.

\subsection{Multimodal Transformer}

To enable effective interaction between image and text modalities, we adopt a multimodal transformer architecture inspired by the MM-DiT block design from SD3~\citep{esser2024scaling}. Unlike prior DiffTSR approaches~\citep{zhang2024diffusion}, which rely on separate modality-specific models connected through an auxiliary communication module, our framework employs a unified transformer backbone. This unified design allows the model to dynamically attend to different modalities depending on the input, thereby supporting joint optimization of both the super-resolution task (LR-to-HR) and the recognition task (LR-to-text). A detailed description can be find at Appendix~\ref{app:joint_attention}

\subsection{Training}
\label{sec:training}
To enable synchronized image and text generation at inference time, we optimize the continuous image process and the discrete text process jointly. Given a training triple $(\mathbf{x}^{\text{hr}}, \mathbf{x}^{\text{lr}}, \mathbf{x}^{\text{txt}})$, the HR image is encoded by a variational autoencoder~\citep{9578911} into the latent space used by the flow-matching branch, while the text sequence is embedded into discrete tokens. For simplicity, we keep the notation $\mathbf{x}^{\text{hr}}$ and $\mathbf{x}^{\text{lr}}$ for the corresponding latent variables in the equations below.

Let $\mathbf{v}_\theta$ denote the image-velocity head and $p_\theta$ denote the text-denoising head of the shared multimodal transformer. We sample
\[
\mathbf{x}^{\text{hr}}_{t} = (1-t)\mathbf{x}^{\text{hr}} + t\bm{\epsilon}, \qquad \bm{\epsilon}\sim\mathcal{N}(0,I),
\]
and corrupt the text with the absorbing process in Equation~\ref{eq:mdlm_xt} to obtain $\mathbf{x}^{\text{txt}}_t$. The corresponding image target velocity is $\mathbf{u}_t=\bm{\epsilon}-\mathbf{x}^{\text{hr}}$.

We first optimize modality-specific conditional objectives. For image generation, the model predicts the HR velocity conditioned on the LR image and the clean text:
\begin{align}
\mathcal{L}_{\text{IMG}} = \mathbb{E}_{t,q} \left\| \mathbf{v}_{\theta}\left( \mathbf{x}^{\text{hr}}_{t}, t, \mathbf{c}^{\text{img}} \right) - \mathbf{u}_t \right\|_2^2,
\end{align}
where $\mathbf{c}^{\text{img}}=\{\mathbf{x}^{\text{lr}}, \mathbf{x}^{\text{txt}}\}$. For text prediction, the model reconstructs the clean text from corrupted tokens conditioned on the LR image and the clean HR image:
\begin{align}
\mathcal{L}_{\text{TXT}} = \mathbb{E}_{q^{(\text{txt})}} \left[
- \frac{1}{K} \sum_{i=1}^{K} \log p_\theta\left(\mathbf{x}^{\text{txt}} \mid \mathbf{x}^{\text{txt}}_{t_i}, t_i, \mathbf{c}^{\text{txt}} \right)
\right],
\end{align}
where $\mathbf{c}^{\text{txt}}=\{\mathbf{x}^{\text{lr}}, \mathbf{x}^{\text{hr}}\}$. Following~\citet{Li_2025_CVPR}, we approximate the continuous-time objective with antithetic sampling~\citep{NEURIPS2021_b578f2a5} over $K$ timesteps $t_i$ uniformly covering $(\delta,1]$, where $\delta$ is a small constant for numerical stability.

To strengthen joint modeling, we further corrupt both modalities with the same timestep and train the model to recover them simultaneously. This synchronized corruption is important because it exposes the transformer to partially observed image and text states at matched noise levels, forcing the two branches to cooperate rather than behave like loosely coupled auxiliaries:
\begin{align}
\mathcal{L}_{\text{Joint}} = \mathbb{E}_{t,q^{(\text{hr,lr,txt})}} \Big[
&\left\| \mathbf{v}_{\theta}\left( \mathbf{x}^{\text{hr}}_{t}, t, \tilde{\mathbf{c}}^{\text{img}} \right) - \mathbf{u}_t \right\|_2^2 \nonumber\\
&- \log p_\theta\left(\mathbf{x}^{\text{txt}} \mid \mathbf{x}^{\text{txt}}_{t}, t, \tilde{\mathbf{c}}^{\text{txt}} \right)
\Big],
\end{align}
where $\tilde{\mathbf{c}}^{\text{img}}=\{\mathbf{x}^{\text{lr}}, \mathbf{x}^{\text{txt}}_t\}$ and $\tilde{\mathbf{c}}^{\text{txt}}=\{\mathbf{x}^{\text{lr}}, \mathbf{x}^{\text{hr}}_t\}$. The overall training objective is
\begin{equation}
\mathcal{L}_{\text{Overall}} = \mathcal{L}_{\text{IMG}} + \mathcal{L}_{\text{TXT}} + \mathcal{L}_{\text{Joint}}.
\end{equation}

\subsection{Model-Guided Training via Classifier-Free Guidance}

To improve sample quality and controllability at inference time, we incorporate model guidance (MG) directly into training, following \citet{tang2025diffusion}. We apply this idea to the image-flow target in Equation~\ref{eq:cfm_loss}. Let $\mathbf{v}_{\theta}^{\text{ema}}\left( \mathbf{x}_{t}, t, \bm{c} \right)$ be the conditioned velocity predicted by the EMA teacher model and $\mathbf{v}_{\theta}^{\text{ema}}\left( \mathbf{x}_{t}, t, \varnothing \right)$ be the corresponding unconditional prediction. The flow-matching objective is then reformulated as
\begin{equation}
    \mathcal{L}_{\text{CFM-MG}} = \mathbb{E}_{t, q(\mathbf{x}_0), q(\mathbf{x}_1, \mathbf{c})} \left\| \mathbf{v}_\theta(\mathbf{x}_{t}, t, \mathbf{c}) - \bm{u}'_t \right\|^2
\end{equation}
where the rectified target $\bm{u}'_t$ is defined as:
\begin{equation}
    \bm{u}'_t(\mathbf{x}_{t}, t,\bm{c}) = \bm{u}_t + w \cdot \left( \text{sg}\left(\mathbf{v}^{\text{ema}}_\theta(\mathbf{x}_{t}, t,\bm{c})\right) - \mathbf{v}^{\text{ema}}_\theta(\mathbf{x}_{t}, t, \varnothing) \right),
    \label{eq:mg_target}
\end{equation}
where $w$ denotes the guidance scale factor and $\text{sg}(\cdot)$ is the stop-gradient operator. Using the EMA teacher stabilizes the target prediction throughout training. During optimization, the condition vector $\bm{c}$ is randomly replaced with $\varnothing$ with probability $\psi$ so that the network learns both conditional and unconditional predictions.

Applying this rectified target to the image-only and joint image losses gives
\begin{equation}
\begin{aligned}
    &\mathcal{L}_{\text{IMG-MG}} =   \mathbb{E}_{t,q^{(\text{img})}} \\
    & \left\| \mathbf{v}_{\theta}\left( \mathbf{x}^{\text{hr}}_{t}, t, \mathbf{c}^{\text{img}} \right) - \bm{u}'_t(\mathbf{x}^{\text{hr}}_{t}, t,\mathbf{c}^{\text{img}}) \right\|_2^2, \\
    &\mathcal{L}_{\text{Joint-MG}} = 
    \mathbb{E}_{t,q^{(\text{img,txt})}} \Big[
    \left\| \mathbf{v}_{\theta}\left( \mathbf{x}^{\text{hr}}_{t}, t, \tilde{\mathbf{c}}^{\text{img}} \right) - \bm{u}'_t(\mathbf{x}^{\text{hr}}_{t}, t,\tilde{\mathbf{c}}^{\text{img}}) \right\|_2^2 \\
    &\hspace{32mm} - \log p_\theta\left(\mathbf{x}^{\text{txt}} \mid \mathbf{x}^{\text{txt}}_{t}, t, \tilde{\mathbf{c}}^{\text{txt}} \right)\Big].
\end{aligned}
\end{equation}
Our final training objective is then defined as:
\begin{equation}
\mathcal{L}_{\text{Overall}} = \mathcal{L}_{\text{IMG-MG}} + \mathcal{L}_{\text{TXT}} + \mathcal{L}_{\text{Joint-MG}}.
\end{equation}

\subsection{Joint inference}

At inference time, given a low-resolution image $\mathbf{x}^{\text{lr}}$, we initialize the image branch from Gaussian noise $\mathbf{x}^{\text{hr}}_1$ and the text branch from a fully masked sequence $\mathbf{x}^{\text{txt}}_1$. The shared transformer then updates both modalities synchronously, so the current text hypothesis can immediately influence image refinement and vice versa. For a step from $t$ to $s$ ($0<s<t<1$), the model predicts
\begin{align}
\hat{\mathbf{u}}_t &= \mathbf{v}_{\theta}\left(\mathbf{x}^{\text{hr}}_{t}, t, \{\mathbf{x}^{\text{lr}}, \mathbf{x}^{\text{txt}}_t\}\right), \\
\mathbf{p}_t &= p_\theta\left(\cdot \mid \mathbf{x}^{\text{txt}}_t, t, \{\mathbf{x}^{\text{lr}}, \mathbf{x}^{\text{hr}}_t\}\right).
\end{align}
The image latent is updated with an Euler step,
\begin{equation}
\mathbf{x}^{\text{hr}}_{s} = \mathbf{x}^{\text{hr}}_{t} - (t-s)\hat{\mathbf{u}}_t,
\end{equation}
while the text latent is updated by the reverse absorbing-state transition parameterized by $\mathbf{p}_t$, with only masked positions being resampled and already generated tokens kept fixed. After obtaining the final image latent $\mathbf{x}^{\text{hr}}_0$, we decode it back to RGB space using the VAE decoder. Detailed pseudocode for training and sampling is provided in Appendix~\ref{app:pseudo_code}.

%% file: sec/experiment_new.tex
\begin{table*}[ht]
    \centering
    \resizebox{1.0\hsize}{!}{
    \begin{tabular}{c|ccccc|ccccc}
    \toprule
         \multirow{2}{*}{\bf Method} & \multicolumn{5}{c|}{$\times$\textbf{2}} & \multicolumn{5}{c}{$\times$\textbf{4}} \\ \cline{2-11}
        &\bf PSNR$\uparrow$& \bf LPIPS$\downarrow$& \bf FID$\downarrow$& \bf ACC$\uparrow$& \bf NED$\uparrow$ &\bf PSNR$\uparrow$& \bf LPIPS$\downarrow$& \bf FID$\downarrow$& \bf ACC$\uparrow$& \bf NED$\uparrow$\\
        \midrule
        ESRGAN~\citep{wang2018ESRGAN} & 24.09 & 0.3046 & 15.06 & 67.08\% & 83.79\% & 22.18 & 0.3986 & 18.25 & 43.69\% & 62.15\% \\
        MSRResNet~\citep{9022144} & 28.03 & 0.3030 & 30.47 & 68.94\% & 85.37\% & 24.52 & 0.4029 & 50.60 & 49.04\% & 67.96\% \\
        SwinIR~\citep{liang2021swinir} & 28.39 & 0.2983 & 31.54 & 70.01\% & 86.10\% & 24.73 & 0.3957 & 50.89 & 50.09\% & 68.93\% \\
        SRFormer~\citep{zhou2023srformer} & \textbf{28.89} & 0.2829 & 26.93 & 70.93\% & 86.91\% & \textbf{25.05} & 0.3801 & 46.23 & 51.83\% & 70.70\% \\ \hline
        MARCONet~\citep{li2023learning} & 23.14 & 0.4518 & 88.06 & 57.93\% & 76.43\% & 21.44 & 0.4941 & 99.64 & 41.30\% & 59.86\% \\
        DiffTSR~\citep{zhang2024diffusion} & 23.21 & 0.3304 & 18.57 & 63.51\% & 80.99\% & 20.62 & 0.3952 & 22.24 & 44.87\% & 63.20\% \\
        MARCONet++~\citep{li2025enhanced} & 24.02 & 0.4281 & 64.27 & 59.34\% & 77.49\% & 22.06 & 0.5072 & 86.18 & 39.92\% & 58.30\% \\
        \hline
        \cc{\textbf{DualTSR (Ours)}} & \cc{22.43} & \cc{\textbf{0.2682}} & \cc{\textbf{8.73}} & \cc{\textbf{73.23\%}} & \cc{\textbf{88.50\%}} & \cc{20.54} & \cc{\textbf{0.3292}} & \cc{\textbf{16.42}} & \cc{\textbf{57.65\%}} & \cc{\textbf{76.64\%}} \\
        \bottomrule
    \end{tabular}}
    \caption{\textbf{Quantitative comparison on the synthetic CTR-TSR}. Best in \textbf{bold}.}
    \label{tab:ctr_results}
\end{table*}

\begin{table*}[ht]
    \centering
    \resizebox{1.0\hsize}{!}{
    \begin{tabular}{c|ccccc|ccccc}
    \toprule
         \multirow{2}{*}{\bf Method} & \multicolumn{5}{c|}{$\times \textbf{2}$} & \multicolumn{5}{c}{$\times \textbf{4}$} \\ \cline{2-11}
        &\bf PSNR$\uparrow$& \bf LPIPS$\downarrow$& \bf FID$\downarrow$& \bf ACC$\uparrow$& \bf NED$\uparrow$ &\bf PSNR$\uparrow$& \bf LPIPS$\downarrow$& \bf FID$\downarrow$& \bf ACC$\uparrow$& \bf NED$\uparrow$\\
        \midrule
        ESRGAN~\citep{wang2018ESRGAN} & 20.08 & 0.3833 & 103.50 & 58.80\% & 86.30\% & 19.97 & 0.3721 & 87.18 & 58.50\%& 87.65\% \\
        MSRResNet~\citep{9022144} & 19.25 & \textbf{0.3033} & 43.82 & 62.80\% & 89.08\% & 19.28 & 0.3590 & 60.81 & 58.00\%& 86.87\% \\
        SwinIR~\citep{liang2021swinir} & 20.24 & 0.3810 & 87.98 & 59.30\% & 86.44\% & 20.23 & \textbf{0.3271} & 56.77 & 62.10\%& 87.55\% \\
        SRFormer~\citep{zhou2023srformer} & \textbf{20.85} & 0.3081 & 50.81 & 62.90\% & 89.47\% & 20.29 & 0.3732 & 83.47 & 57.50\%& 85.79\% \\
        \hline
        MARCONet~\citep{li2023learning} & 18.93 & 0.4144 & 87.54 & 60.58\% & 87.11\% & 19.09 & 0.4189 & 92.14 & 61.52\%& 88.39\% \\
        DiffTSR~\citep{zhang2024diffusion} & 18.95 & 0.3179 & 38.13 & 62.60\% & 88.13\% & 18.09 & 0.3382 & 41.13 & 58.00\%& 84.44\% \\
        MARCONet++~\citep{li2025enhanced} & 19.38 & 0.3839 & 75.47 & 61.20\% & 87.47\% & 19.48 & 0.3953 & 81.72 & 59.40\%& 87.76\% \\
        \hline
        \cc{\textbf{DualTSR (Ours)}} & \cc{18.94} & \cc{0.3133} & \cc{\textbf{35.95}} & \cc{\textbf{63.80\%}} & \cc{\textbf{89.97\%}} & \cc{18.85} & \cc{0.3277} & \cc{\textbf{40.78}} & \cc{\textbf{62.20\%}} & \cc{\textbf{88.49\%}} \\
        \bottomrule
    \end{tabular}}
    \caption{\textbf{Quantitative comparison on RealCE}. Best in \textbf{bold}.}
    \label{tab:realce_results}
\end{table*}

\section{Experiments}
\label{sec:experiments}

\subsection{Experimental Setup}

\textbf{Datasets.}
We evaluate on two benchmarks.
First, we construct a Chinese scene text super-resolution dataset, denoted as \textbf{CTR-TSR}, by strictly following the construction pipeline described in DiffTSR~\citep{zhang2024diffusion}: we create training and testing splits from CTR~\citep{yu2021benchmarking} using the same filtering criteria, canonical HR resizing, and blind degradation recipe. Because the exact DiffTSR split is not publicly released, we re-implemented the pipeline per their description; the resulting CTR-TSR contains 64{,}139 training and 8{,}690 testing images. Detailed construction procedures are in Appendix~\ref{app:ctr-tsr-construction}.
Second, we evaluate on the \textbf{RealCE} benchmark~\citep{ma2023benchmark}, which contains real-world Chinese text images. Following DiffTSR~\citep{zhang2024diffusion}, which also notes that RealCE contains samples with partial annotations, inaccurate localization, and severe LR--HR misalignment, we report paired metrics on a curated subset of 300 clean, well-aligned LR--HR pairs. Since the exact subset used in DiffTSR is not publicly released, we construct our own subset according to the same considerations to ensure reliable evaluation.

% Second, we evaluate on the \textbf{RealCE} benchmark~\citep{ma2023benchmark}, which contains real-world Chinese text images. Since many RealCE samples exhibit partial annotations, inaccurate localization, or severe LR--HR misalignment, we report paired metrics on a curated subset of 300 clean, well-aligned LR--HR pairs to ensure measurement reliability.

\noindent\textbf{Evaluation Metrics.}
For both CTR-TSR and RealCE, we report PSNR (↑), LPIPS (↓), FID (↓), and two text fidelity metrics: ACC (↑) and NED (↑)~\citep{zhang2024diffusion}.
ACC is the exact-match recognition accuracy (word-level); NED measures character level similarity by normalizing the Levenshtein distance~\citep{zhang2024diffusion}.
To avoid evaluator drift, we use a fixed pretrained OCR recognizer (TransOCR from CTR~\citep{yu2021benchmarking}) for all methods unless otherwise noted; computation details for the text metric are in Appendix~\ref{app:text-metrics}.

\noindent\textbf{Degradation and Resolutions.}
Unless specified, HR images are resized to $128\times512$ (text-line layout preserved). LR images are synthesized by a blind, stochastic pipeline (blur + noise + compression + downsampling, with random ordering/magnitudes) adapted from BSRGAN~\cite{zhang2021designing} and Real-ESRGAN~\cite{wang2021real}; full recipe and randomization ranges are summarized in Appendix~\ref{app:ctr-tsr-construction}.

\noindent\textbf{Training Configuration.}
Our model is implemented in PyTorch and trained on 4$\times$NVIDIA A100 (40 GB) GPUs.
We use AdamW~\citep{loshchilov2018decoupled} with learning rate $1\times 10^{-4}$, cosine decay, and weight decay $0.05$.
For CTR-TSR, we train for 700k iterations with batch size 32 (global).
By default, we adopt $K=8$ discrete-text timesteps for antithetic sampling in the text diffusion loss and a 4-step ODE sampler for joint inference unless otherwise stated. Unless noted otherwise, we set the guidance scale factor $w$ in Equation~\ref{eq:mg_target} to 1.0, which provides the best overall trade-off in Table~\ref{tab:cfg_ablation}. Code, dataset construction scripts, and pretrained checkpoints will be released.

\noindent\textbf{Baselines.} We compare with general SR methods (ESRGAN~\citep{wang2018ESRGAN}, MSRResNet~\citep{9022144}, SwinIR~\citep{liang2021swinir}, SRFormer~\citep{zhou2023srformer}) and text-specific approaches (MARCONet~\citep{li2023learning}, MARCONet++~\citep{li2025enhanced}, DiffTSR~\citep{zhang2024diffusion}).
General SR baselines are retrained on CTR-TSR-Train. For text-specific methods, we use official checkpoints: our CTR-TSR dataset strictly follows the construction pipeline described in DiffTSR~\citep{zhang2024diffusion} (same source corpus, filtering criteria, and degradation recipe), so the evaluation is comparable even without retraining. MARCONet/MARCONet++ ship their own synthetic pipelines and are designed to generalize to unseen degradations, making official-checkpoint evaluation the standard practice. Full reproduction details are in Appendix~\ref{app:repro}.

\begin{figure*}[!ht]
  \begin{center}
    \includegraphics[width=\linewidth]{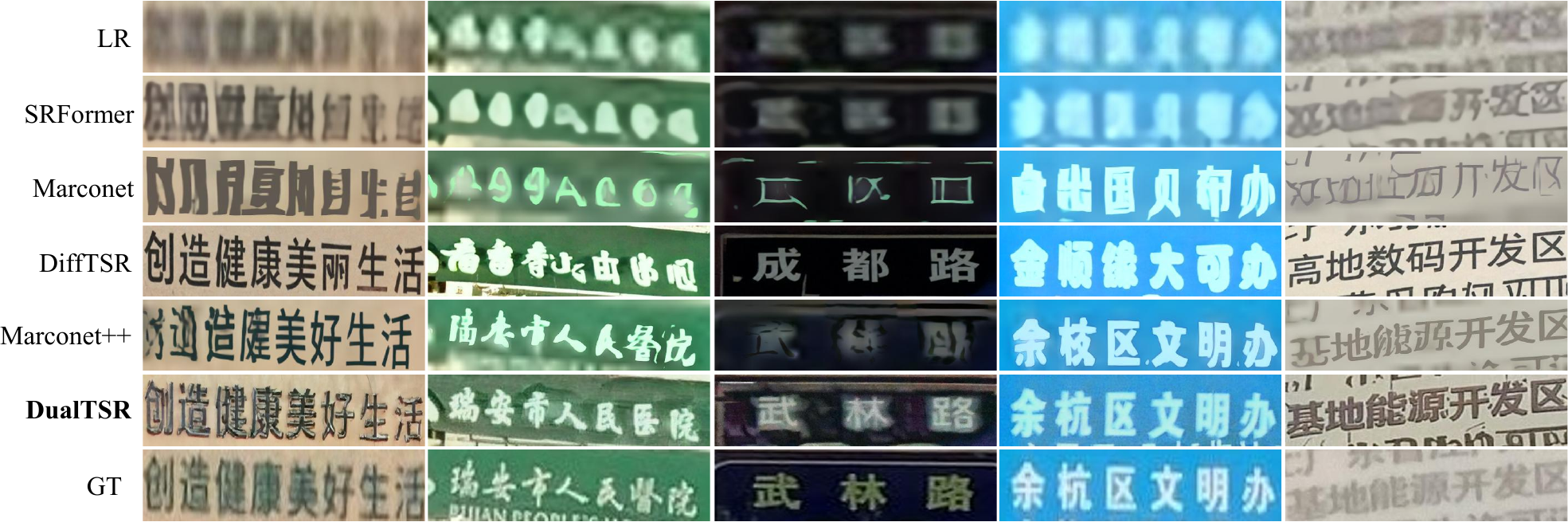}
  \end{center}
  \caption{\textbf{Qualitative comparison for RealCE with different methods on $\times$4 scale.}
  The comparison methods include ESRGAN~\citep{wang2018ESRGAN}, MSRResNet~\citep{9022144}, SwinIR~\citep{liang2021swinir}, SRFormer~\citep{zhou2023srformer}, MARCONet~\citep{li2023learning}, MARCONet++~\citep{li2025enhanced}, DiffTSR~\citep{zhang2024diffusion} and our method.}
  \label{fig:realce_X4}
\end{figure*}

\subsection{Main Results}

\noindent\textbf{Quantitative.}
Tables~\ref{tab:ctr_results} and~\ref{tab:realce_results} report performance on the synthetic CTR-TSR benchmark and the real-world RealCE benchmark. On CTR-TSR, generic SR models (ESRGAN, MSRResNet, SwinIR, SRFormer) achieve strong PSNR but poor LPIPS/FID and low ACC/NED at both scales, confirming that pixel-level reconstruction alone is insufficient for recovering structured text.
MARCONet and MARCONet++ introduce text-aware supervision, yet their compositing-based training often disrupts scene consistency and limits perceptual performance. DiffTSR improves FID through its dual diffusion branches, but the separation of text and image pathways still leaves room for semantic errors. DualTSR achieves the best FID, LPIPS, ACC, and NED on CTR-TSR across both $\times2$ and $\times4$ settings. As expected for a generative model, PSNR is lower than regression-based baselines; this reflects the well-known perception-distortion trade-off~\cite{xue2023burst}, where perceptual realism and pixel-wise fidelity are inherently at odds.

On the curated RealCE subset, real-world degradations amplify the weaknesses of prior approaches: generic SR models preserve coarse textures but often fail to reconstruct valid glyphs, while text-specific baselines show limited robustness beyond their synthetic training assumptions. DualTSR achieves the best ACC and NED at both scales together with competitive perceptual quality, demonstrating that the unified backbone transfers effectively to real-world imagery.

% Overall, the two benchmarks highlight complementary aspects of the task: CTR-TSR emphasizes semantic correctness under controlled degradations, whereas RealCE stresses robustness to real imaging artifacts. {DualTSR is the only method that performs consistently well across both}, validating the effectiveness of unified visual–text generation.

\noindent\textbf{Qualitative.}
% Figures~\ref{fig:ctr_test_X4} and~\ref{fig:realce_X4} provide $\times4$ visual comparisons on CTR-TSR and RealCE.
Figures~\ref{fig:realce_X4} provide $\times4$ visual comparisons on RealCE. We provide more visual comparisons on both CTR-TSR and RealCE at Appendix~\ref{app:more_qualit}.
Three recurring patterns emerge.
\emph{(i) Structural distortion.} Generic SR models (ESRGAN, MSRResNet, SwinIR, SRFormer) tend to oversmooth strokes or introduce spurious edges, reflecting their lack of text-specific supervision.
\emph{(ii) Style inconsistency.} MARCONet and MARCONet++ sometimes produce correct character identities but generate glyphs that appear stylistically pasted onto the background, breaking scene coherence.
\emph{(iii) Semantic errors.} DiffTSR yields sharper characters than MARCONet variants, yet its separately trained text and image branches can produce semantically incorrect glyphs.
DualTSR mitigates all three failure modes: the shared backbone preserves local appearance cues (font, color, background) while the internal text hypothesis guides glyph structure, producing characters that are both semantically correct and visually integrated with the scene. Failure cases remain under severe corruption when color or font cues are largely missing; additional $\times2$ results are in Appendix~\ref{app:more_qualit}.

\subsection{Efficiency Analysis}
\label{sec:efficiency}

We evaluate end-to-end inference efficiency on CTR-TSR $\times4$
using a single NVIDIA A100, batch size 1, and an output resolution
of $128\times512$. Parameters and peak memory include all modules
required at inference. As shown in Table~\ref{tab:efficiency},
DualTSR requires only four network function evaluations (NFEs),
completing inference in 132\,ms with 1.2\,GB peak memory.
Compared with DiffTSR, DualTSR uses $150\times$ fewer NFEs, runs
$100.9\times$ faster, and reduces the parameter count and memory
footprint by $6.1\times$ and $4.5\times$, respectively, while
improving ACC by 12.78 percentage points. Moreover, its latency is
within the range of feed-forward SR models, demonstrating that the
unified image--text backbone substantially reduces the cost of
generative STISR while retaining strong text fidelity.

\begin{table}[t]
\centering
\resizebox{\linewidth}{!}{%
\begin{tabular}{lccccc}
\toprule
\textbf{Method}
& \textbf{Params}
& \textbf{NFE}
& \textbf{Latency}$\downarrow$
& \textbf{Peak Mem.}$\downarrow$
& \textbf{ACC}$\uparrow$ \\
\midrule
DiffTSR
& 1,231\,M
& $200\times3$
& 13,316\,ms
& 5.4\,GB
& 44.87\% \\
SRFormer
& 20\,M
& 1
& 70\,ms
& 0.3\,GB
& 51.83\% \\
\textbf{DualTSR (Ours)}
& 203\,M
& 4
& 132\,ms
& 1.2\,GB
& \textbf{57.65\%} \\
\bottomrule
\end{tabular}%
}
\caption{\textbf{Efficiency comparison on CTR-TSR $\times4$.}
Runtime is measured on a single NVIDIA A100 with batch size 1 and
an output resolution of $128\times512$. NFE denotes the number of
network function evaluations.}
\label{tab:efficiency}
\end{table}

\subsection{Internal Text-Branch Evaluation}
\label{sec:internal_text}

We directly evaluate the text predictions produced by DualTSR on CTR-TSR $\times4$ against the ground-truth transcriptions. Table~\ref{tab:internal_text} compares its internal text branch with that of DiffTSR and reports OCR results obtained by applying TransOCR to the restored images. 
DualTSR substantially outperforms DiffTSR in both internal prediction and OCR-based evaluation. Its smaller gap between internal and OCR accuracy further suggests better alignment between predicted text semantics and restored glyphs.

\begin{table}[t]
\centering
\resizebox{\linewidth}{!}{
\begin{tabular}{lcccc}
\toprule
\multirow{2}{*}{\textbf{Method}} 
& \multicolumn{2}{c}{\textbf{Internal}} 
& \multicolumn{2}{c}{\textbf{OCR on SR}} \\
\cmidrule(lr){2-3}\cmidrule(lr){4-5}
& ACC$\uparrow$ & NED$\uparrow$ & ACC$\uparrow$ & NED$\uparrow$ \\
\midrule
DiffTSR~\citep{zhang2024diffusion} 
& 26.09\% & 47.27\% & 44.87\% & 63.20\% \\
\textbf{DualTSR (Ours)} 
& \textbf{45.44\%} & \textbf{69.66\%} 
& \textbf{57.65\%} & \textbf{76.64\%} \\
\bottomrule
\end{tabular}}
\caption{\textbf{Internal text evaluation on CTR-TSR $\times4$.} ``OCR on SR'' denotes TransOCR evaluation of the restored images.}
\label{tab:internal_text}
\end{table}

\begin{table}[t]
    \centering
    \resizebox{\linewidth}{!}{
    \begin{tabular}{lccccc}
        \toprule
        Setting & FID$\downarrow$ & PSNR$\uparrow$ & LPIPS$\downarrow$ & ACC$\uparrow$ & NED$\uparrow$ \\
        \midrule
        (a) $\mathcal{L}_{\text{Joint-MG}}$ & 13.73&19.82&0.3804&49.85\%&68.28\%\\
        (b) $\mathcal{L}_{\text{Joint-MG}} + \mathcal{L}_{\text{TXT}}$ & 13.39&19.79&0.3782&52.72\%& 72.17\%\\
        {(c) $\mathcal{L}_{\text{Joint-MG}} + \mathcal{L}_{\text{TXT}} + \mathcal{L}_{\text{IMG-MG}}$ }  & 9.92&20.12&0.3550&53.71\%&73.60\%\\
        \bottomrule
    \end{tabular}
    }
    \caption{\textbf{Loss ablation on CTR-TSR $\times$4.} All variants are trained for 300k iterations with batch size 128 for efficient comparison.}
    \label{tab:abl_components}
\end{table}

\subsection{Ablation Study}

\paragraph{Loss design ablation study.}
Table~\ref{tab:abl_components} presents an ablation on our loss components. Starting with the baseline {(a) $\mathcal{L}_{\text{Joint-MG}}$}, the model already achieves strong results by jointly corrupting image and text with synchronized timesteps, yielding a balanced performance across both image quality (FID 13.73, LPIPS 0.3804) and text reconstruction (ACC 49.85\%, NED 68.28\%).
Adding the discrete text diffusion loss {(b) $\mathcal{L}_{\text{Joint-MG}} + \mathcal{L}_{\text{TXT}}$} further improves the text metrics substantially (ACC +2.87, NED +3.89) while slightly reducing FID and LPIPS, indicating that explicit supervision on the text pathway stabilizes the shared Transformer and benefits both modalities.

Finally, incorporating the full image–text training objective {(c) $\mathcal{L}_{\text{Joint-MG}} + \mathcal{L}_{\text{TXT}} + \mathcal{L}_{\text{IMG-MG}}$} yields the best overall performance, improving FID from 13.39 to 9.92 and LPIPS from 0.3782 to 0.3550, while also boosting text accuracy to 53.71\%. These gains show that the image MG loss complements the discrete text diffusion: image–level gradient signals regularize the latent Transformer and ensure that text reconstruction aligns with visually plausible high-resolution structures. These results show that Joint-MG effectively couples image and text generation, while the modality-specific losses further refine their respective outputs. We also provide a effect of sampling on Appendix~\ref{app:sampling_step}.

% Overall, the ablation confirms the \emph{Joint-MG} mechanism is essential for coupling the two modalities, while the additional image/text losses strengthen modality-specific reconstruction. The best results are achieved when all losses operate together, demonstrating that synchronized corruption paired with both continuous (image) and discrete (text) objectives leads to the most coherent multimodal generation.

% \paragraph{Loss design ablation.}
% Table~\ref{tab:abl_components} evaluates each loss component. The Joint-MG baseline achieves balanced image and text reconstruction. Adding $\mathcal{L}_{\text{TXT}}$ improves ACC and NED by 2.87 and 3.89 points, respectively, while also slightly improving FID and LPIPS. Further introducing $\mathcal{L}_{\text{IMG-MG}}$ yields the best overall performance, reducing FID from 13.39 to 9.92 and LPIPS from 0.3782 to 0.3550, while increasing ACC to 53.71\%. These results show that Joint-MG effectively couples image and text generation, while the modality-specific losses further refine their respective outputs.

\begin{figure}[!t]
  \begin{center}
    \includegraphics[width=1\linewidth]{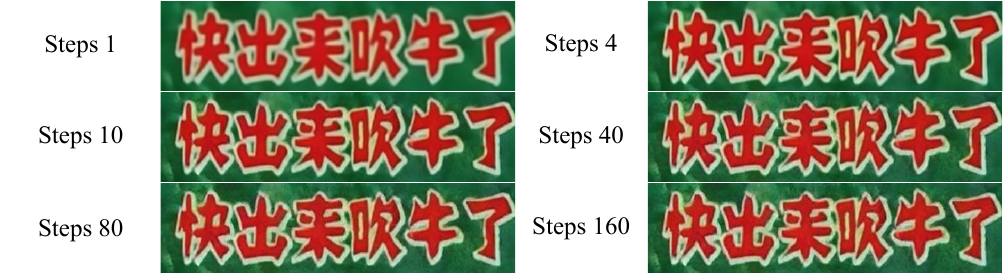}
  \end{center}
  \caption{\textbf{Visual examples across sampling steps.} More steps improve texture smoothness but can blur fine character strokes.}
  \label{fig:step_visual}
\end{figure}

\begin{table}[!t]
\centering
\caption{\textbf{Effect of guidance scale $w$ on CTR-TSR $\times$4.} Increasing $w$ worsens FID and eventually harms NED.}
\resizebox{\linewidth}{!}{
\begin{tabular}{c|ccccc}
\toprule
\textbf{Guidance scale $w$} & 1 & 1.45 & 2 & 4 & 6 \\
\midrule
\textbf{FID$\downarrow$} & 16.10 & 19.59 & 24.19 & 35.35 & 39.49 \\
\textbf{NED$\uparrow$} & 76.66\% & 76.71\% & 75.95\% & 73.09\% & 72.20\% \\
\bottomrule
\end{tabular}}
\label{tab:cfg_ablation}
\end{table}

\paragraph{Effect of guidance scale.}
Table~\ref{tab:cfg_ablation} shows the impact of the guidance scale $w$ (Equation~\ref{eq:mg_target}) on our model. In contrast to typical image synthesis, where stronger guidance often improves perceptual sharpness~\cite{ho2021classifierfree}, we find that increasing $w$ consistently worsens FID, while NED remains nearly unchanged at $w{=}1.45$ and then declines as guidance becomes stronger. Larger $w$ values push the model toward overly confident predictions that deviate from the true data distribution, leading to unnatural textures and distorted character structures. We therefore use $w{=}1.0$, which offers the best overall trade-off. We hypothesize that strong guidance amplifies high-frequency hallucinations during sampling, which harms both visual naturalness and the fine-grained stroke patterns essential for accurate text reconstruction.

%% file: sec/5_conclusion.tex
\section{Conclusion}
\label{sec:conclusion}

We introduced \texttt{DualTSR}, a unified and efficient STISR framework that couples continuous image restoration and discrete text prediction within a shared multimodal transformer. By predicting text internally and allowing image and text states to interact throughout generation, DualTSR avoids externally predicted OCR priors and separate image and text generative backbones. It achieves the best FID, LPIPS, ACC, and NED among the compared methods on CTR-TSR, and the best FID, ACC, and NED with competitive LPIPS on the aligned RealCE subset. At $\times4$, compared with DiffTSR, DualTSR improves ACC by 12.78 percentage points while reducing the parameter count from 1.23\,B to 203\,M, and latency from 13.3\,s to 132\,ms. Its internal text branch also substantially outperforms DiffTSR's dedicated text module, supporting shared continuous-discrete generation as a practical effective STISR method.

%% file: sec/6_appendix.tex
\appendix
{
\onecolumn
\centering
\Large
\textbf{Coupled Continuous–Discrete Generation for Scene Text Image Super-Resolution}\\
% \vspace{0.5em}Supplementary Material \\
\vspace{1.0em}
}

\section{CTR-TSR dataset construction}
\label{app:ctr-tsr-construction}

We build the \textbf{CTR-TSR} dataset from the CTR~\citep{yu2021benchmarking} corpus following the preprocessing and synthesis procedure described in DiffTSR~\citep{zhang2024diffusion}. In brief, the construction pipeline is:

\begin{enumerate}
  \item \textbf{Source images.} Start from the scene part of CTR dataset~\citep{yu2021benchmarking} as the pool of high-quality HR text line images.
  \item \textbf{Filtering and selection.} Retain only HR images that satisfy all of the following criteria:
    \begin{itemize}
      \item resolution (longer side) $\geq 64$ pixels;
      \item width-to-height ratio $> 2$ (to focus on text line images);
      \item length of text annotation $\leq 24$ characters.
    \end{itemize}
    After filtering and resizing (next step), this yields 64139 HR images which we denote {CTR-TSR-Train}.
    
  \item \textbf{HR normalization.} Resize each selected HR image to a canonical HR size of $128\times512$ for training (preserving the overall layout of text lines).
  \item \textbf{LR synthesis (degradation).} We generate LR images using a blind, high-complexity degradation pipeline inspired by BSRGAN~\citep{zhang2021designing} and Real-ESRGAN~\citep{wang2021real}. During training, one of the two degradation strategies is randomly selected with equal probability ($p=0.5$) for each sample, and the corresponding LR image is synthesized online.
  \item \textbf{Synthetic test set.} For the synthetic evaluation set ({CTR-TSR-Test}) we select images from the scene CTR test split, apply the same filtering/resizing and the same degradation pipeline as above. The resulting CTR-TSR-Test contains 8690 LR–HR pairs.
\end{enumerate}

All synthesized LR images used for training and synthetic testing are generated with the same stochastic degradation pipeline so that training and evaluation are comparable across methods.

\begin{figure}[h]
  \begin{center}
    \includegraphics[width=0.3\linewidth]{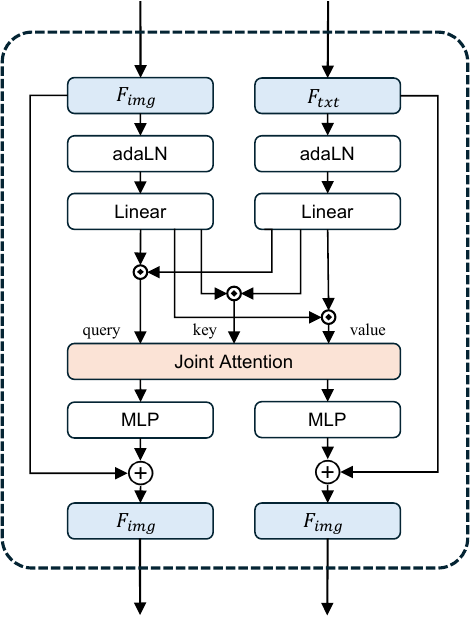}
  \end{center}
  \caption{\textbf{Illustration of the joint attention mechanism.}
Following the MM-DiT block design in SD3~\citep{esser2024scaling}, we incorporate a joint attention module that enables the image latent refinement and text reconstruction processes to operate in parallel.}
  \label{fig:joint_attention}
\end{figure}

\section{Joint attention mechanism}
\label{app:joint_attention}
At the heart of this design is a joint attention mechanism~\citep{esser2024scaling}. As illustrated in Figure~\ref{fig:joint_attention}, latent tokens from the image and text processes are first projected into modality-specific streams. During attention, tokens from both modalities are concatenated to form shared query, key, and value representations and passed through a single self-attention block; the outputs are then split back into image and text streams. This mechanism provides layer-wise multimodal fusion rather than occasional cross-branch communication, enabling visual features to guide text reconstruction and textual cues to influence image restoration at every layer.

\begin{figure}[t]
  \begin{center}
    \includegraphics[width=0.3\linewidth]{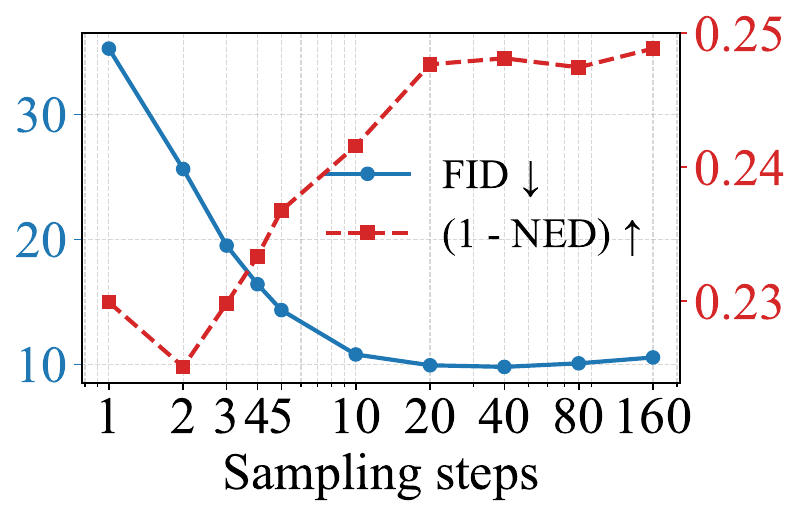}
  \end{center}
  \caption{\textbf{Trade-off between perceptual quality and text fidelity across sampling steps.} 
FID improves with more sampling steps, reaching its best value around \(\sim\!40\), while NED worsens as text strokes become over-smoothed. 
We select \(4\) steps as a balanced operating point.}
  \label{fig:trade_off_sampling_step}
\end{figure}

\section{Effect of sampling steps.}
\label{app:sampling_step}
Figure~\ref{fig:trade_off_sampling_step} illustrates the trade-off between perceptual realism and text fidelity as we vary the {total number of sampling steps} during inference. Increasing the number of steps generally improves FID, with the best perceptual quality achieved around 40 steps, reflecting more effective refinement of textures and global coherence.
In contrast, NED peaks at only 2 steps and degrades steadily as additional refinement is applied, indicating diminished character legibility. We hypothesize that with very few sampling steps, the model focuses on reconstructing coarse structures and high-contrast edges, which are crucial cues for recognizing fine-grained glyph shapes. As the total steps increase, the sampler devotes more iterations to enhancing textural realism and smoothing artifacts (see Fig.~\ref{fig:step_visual}), but this refinement can inadvertently erode subtle, high-frequency stroke patterns—particularly problematic for dense scripts like Chinese where small details distinguish similar characters. Balancing these two factors, we adopt 4 sampling steps as our default setting, which provides a favorable compromise between natural image appearance and accurate text reconstruction.

\section{Reproduction Details for Different Methods}
\label{app:repro}
For MSRResNet~\citep{9022144} and ESRGAN~\citep{wang2018ESRGAN}, we use the BasicSR\footnote{https://github.com/XPixelGroup/BasicSR/tree/master} codebase to train the models on our CTR-TSR-Train dataset with the default settings. For SwinIR~\citep{liang2021swinir}, we train using the official code with the default configuration on CTR-TSR-Train. For SRFormer~\citep{zhou2023srformer}, we likewise use the official code and training settings from the original paper. All four generic SR baselines are retrained on CTR-TSR-Train to ensure a fair comparison under the same data distribution.

For DiffTSR~\citep{zhang2024diffusion}, we use the official checkpoint for direct inference on our test splits. Our CTR-TSR dataset strictly follows the construction pipeline described in DiffTSR (same source corpus, filtering criteria, and degradation recipe), so the evaluation is directly comparable. For MARCONet~\citep{li2023learning} and MARCONet++~\citep{li2025enhanced}, we use the publicly available checkpoints because these models rely on carefully designed data-synthesis pipelines and are intended to generalize to unseen degradations. We observe that MARCONet++ requires detectable text in the LR image; when the model fails to detect text, inference raises an error. In such cases, we use the LR image directly as the model output. This issue is especially common in real-world samples whose text can barely be recognized.

\section{Protocol Note on RealCE}
RealCE was originally released for real-world text image super-resolution, but paired metric evaluation is difficult when adapting it to line-level Chinese STISR because some samples contain partial annotations, inaccurate localization, or severe LR--HR misalignment. We therefore curate a subset of 300 clean LR--HR pairs with verified annotations and accurate alignment. This protocol improves measurement stability and reproducibility; we will release the exact sample list to enable future comparisons under the same protocol.

\section{Text-Related Metrics}
\label{app:text-metrics}

\paragraph{ACC (word accuracy / recognition accuracy).}
ACC is the exact-match recognition accuracy: the fraction of test samples for which the predicted text sequence equals the ground-truth transcription. In all experiments in this paper, we use the pretrained TransOCR recognition model from~\cite{yu2021benchmarking} (as in DiffTSR) to obtain predictions used in ACC calculation. Higher ACC indicates better preservation of textual content in the restored image.

\paragraph{NED (Normalized Edit Distance).}
NED quantifies character-level similarity by normalizing the Levenshtein (edit) distance between the predicted and ground-truth strings. For a single sample with ground truth $g$ and prediction $p$ we compute:
\[
\mathrm{NED}(p,g) = 1 - \frac{\mathrm{ED}(p,g)}{\max(|g|,\,|p|)},
\]
where $\mathrm{ED}(\cdot,\cdot)$ is the Levenshtein distance and $|g|$, and $|p|$ is the length of the ground truth and prediction. We report the dataset mean NED; larger values indicate better (closer) recognition. In CTR-TSR evaluations ACC and NED are computed using the pretrained TransOCR recognizer, as in DiffTSR.

\section{More Qualitative examples}
\label{app:more_qualit}
\subsection{On synthetic CTR-TSR}
\begin{figure}[H]
  \begin{center}
    \includegraphics[width=\linewidth]{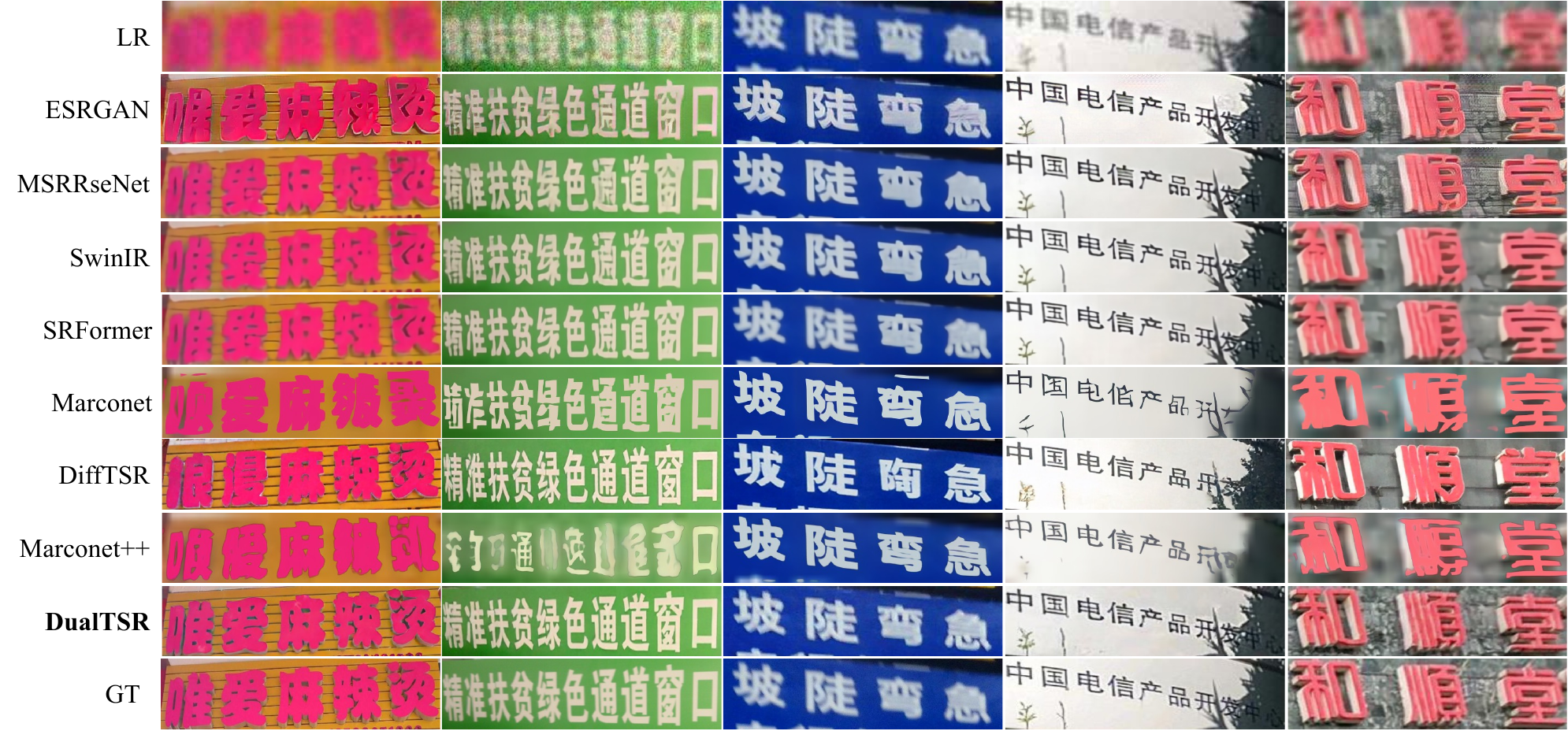}
  \end{center}
  \caption{\textbf{Visual comparison for the synthetic dataset CTR-TSR-Test with different methods on $\times2$ scale.}
  The comparison methods include ESRGAN~\citep{wang2018ESRGAN}, MSRResNet~\citep{9022144}, SwinIR~\citep{liang2021swinir}, SRFormer~\citep{zhou2023srformer}, MARCONet~\citep{li2023learning}, MARCONet++~\citep{li2025enhanced}, DiffTSR~\citep{zhang2024diffusion} and our method.}
  \label{fig:CTR_x2_1}
\end{figure}

\begin{figure}[H]
  \begin{center}
    \includegraphics[width=\linewidth]{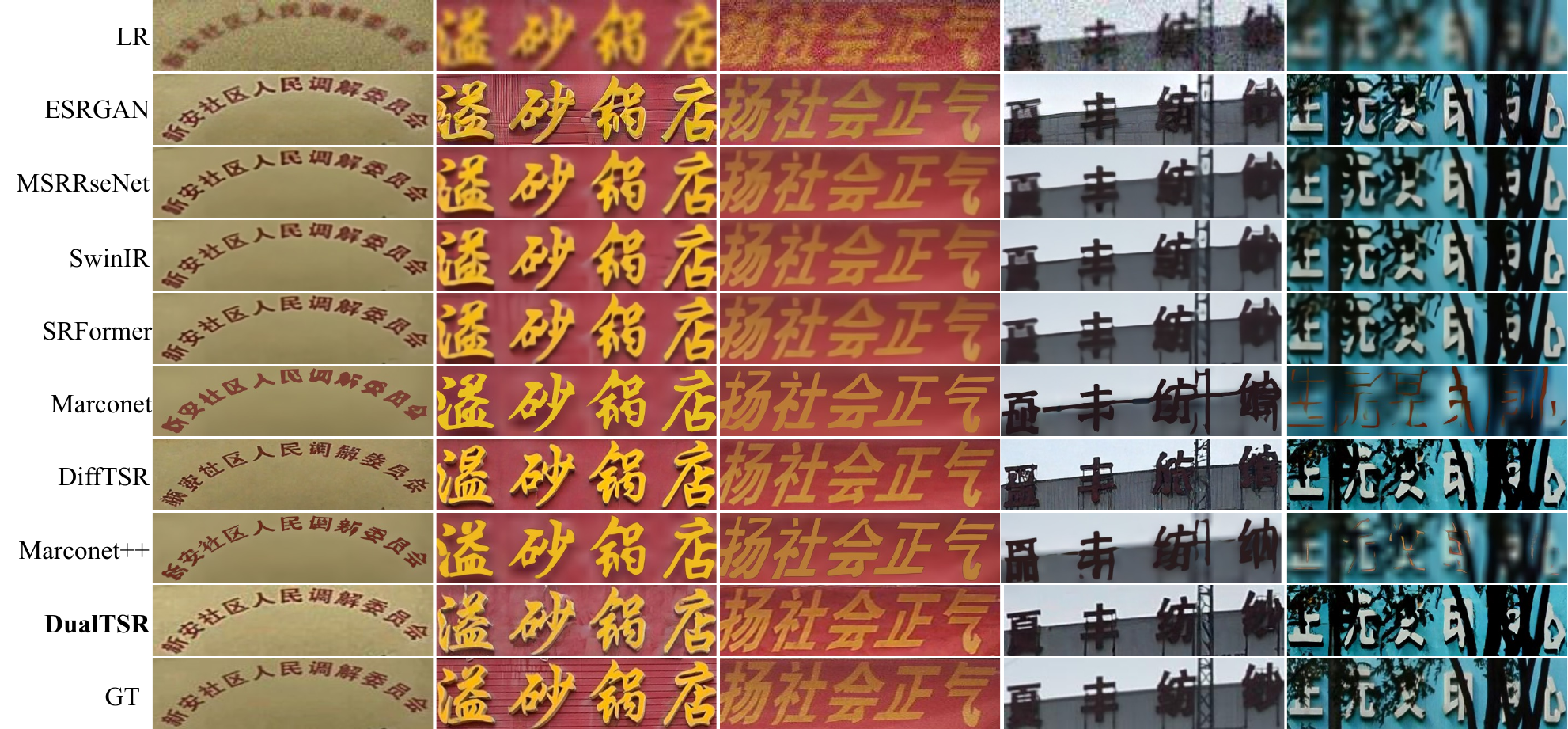}
  \end{center}
  \caption{\textbf{Visual comparison for the synthetic dataset CTR-TSR-Test with different methods on $\times2$ scale.}
  The comparison methods include ESRGAN~\citep{wang2018ESRGAN}, MSRResNet~\citep{9022144}, SwinIR~\citep{liang2021swinir}, SRFormer~\citep{zhou2023srformer}, MARCONet~\citep{li2023learning}, MARCONet++~\citep{li2025enhanced}, DiffTSR~\citep{zhang2024diffusion} and our method.}
  \label{fig:CTR_x2_2}
\end{figure}

\begin{figure}[H]
  \begin{center}
    \includegraphics[width=\linewidth]{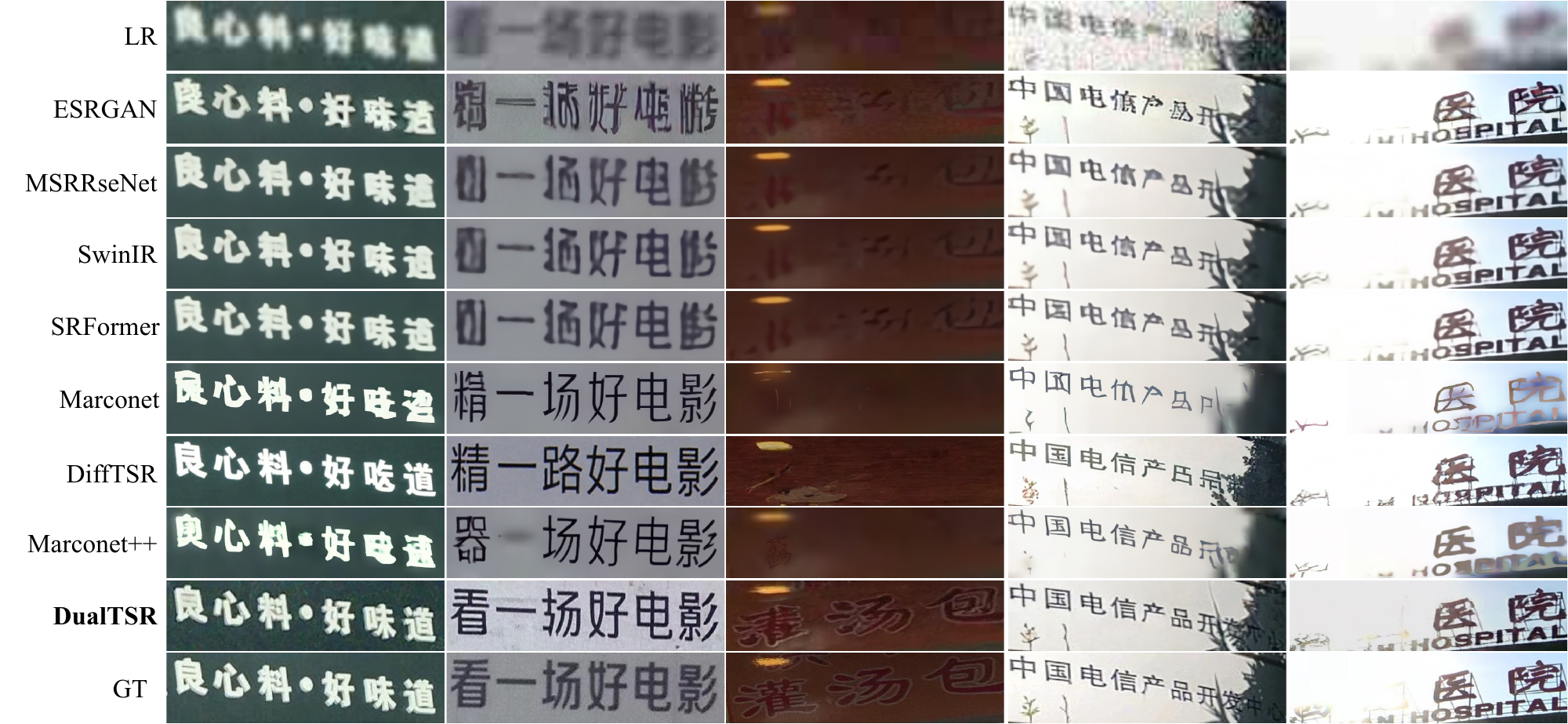}
  \end{center}
  \caption{\textbf{Visual comparison for the synthetic dataset CTR-TSR-Test with different methods on $\times4$ scale.}
  The comparison methods include ESRGAN~\citep{wang2018ESRGAN}, MSRResNet~\citep{9022144}, SwinIR~\citep{liang2021swinir}, SRFormer~\citep{zhou2023srformer}, MARCONet~\citep{li2023learning}, MARCONet++~\citep{li2025enhanced}, DiffTSR~\citep{zhang2024diffusion} and our method.}
  \label{fig:CTR_x4_2}
\end{figure}

For the synthetic CTR-TSR test set, we provide additional $\times2$ results in Figures~\ref{fig:CTR_x2_1} and~\ref{fig:CTR_x2_2}, and additional $\times4$ results in Figure~\ref{fig:CTR_x4_2}.

\begin{figure*}[!ht]
  \begin{center}
    \includegraphics[width=\linewidth]{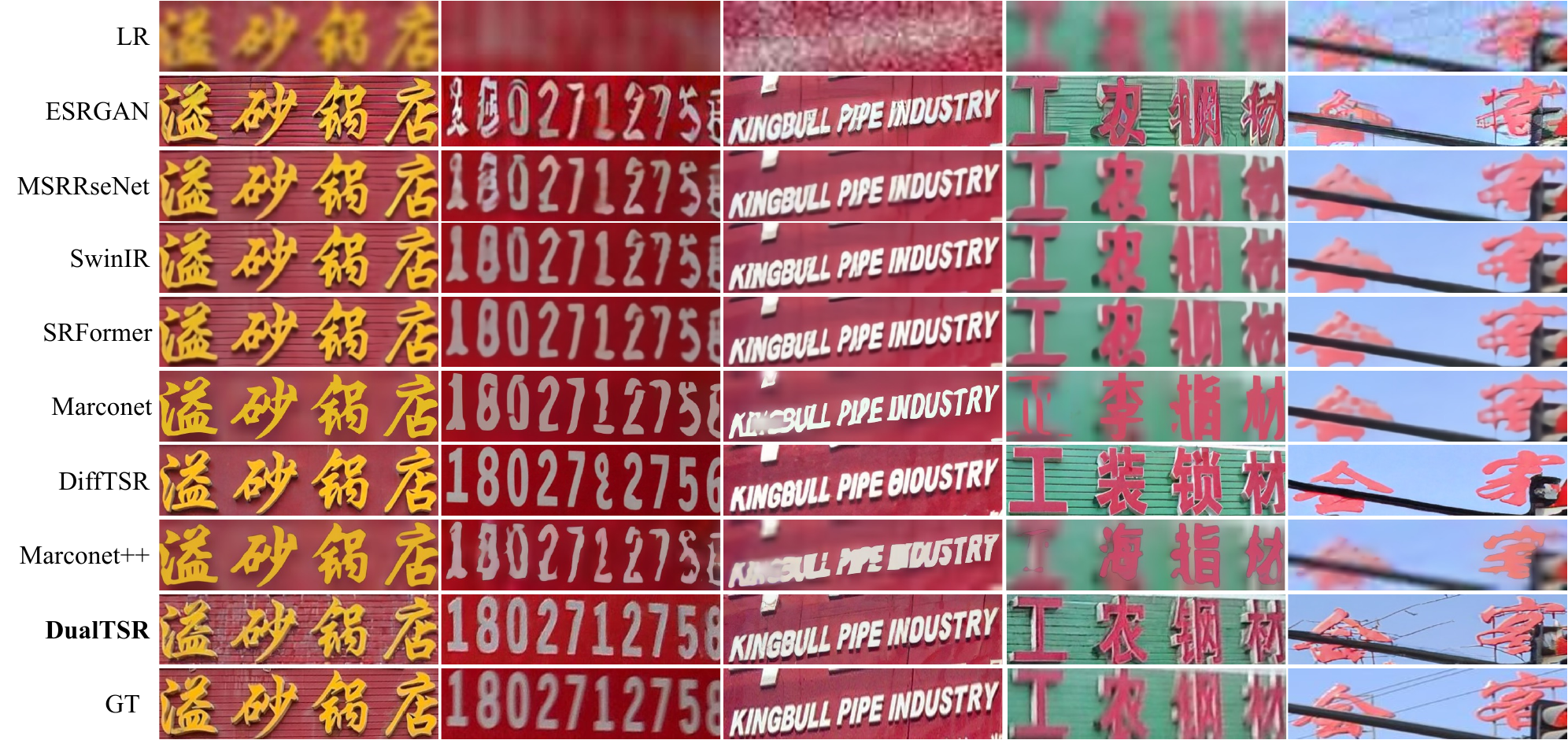}
  \end{center}
  \caption{\textbf{Qualitative comparison for the synthetic dataset CTR-TSR-Test with different methods on $\times$4 scale.}
  The comparison methods include ESRGAN~\citep{wang2018ESRGAN}, MSRResNet~\citep{9022144}, SwinIR~\citep{liang2021swinir}, SRFormer~\citep{zhou2023srformer}, MARCONet~\citep{li2023learning}, MARCONet++~\citep{li2025enhanced}, DiffTSR~\citep{zhang2024diffusion} and our method.}
  \label{fig:ctr_test_X4}
\end{figure*}


\subsection{On real-world dataset RealCE}
\begin{figure}[H]
  \begin{center}
    \includegraphics[width=\linewidth]{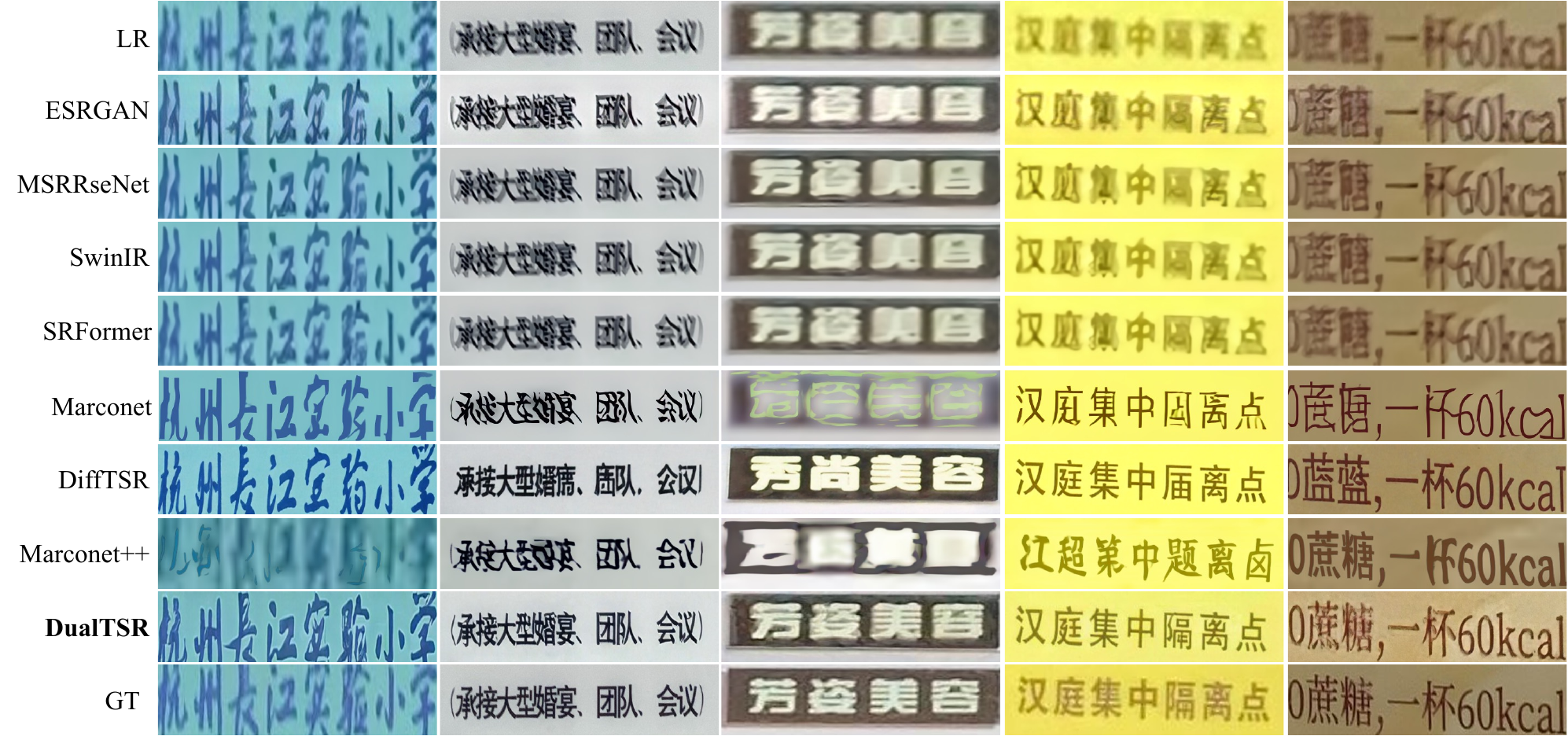}
  \end{center}
  \caption{\textbf{Visual comparison for the real-world dataset RealCE with different methods on $\times2$ scale.}
  The comparison methods include ESRGAN~\citep{wang2018ESRGAN}, MSRResNet~\citep{9022144}, SwinIR~\citep{liang2021swinir}, SRFormer~\citep{zhou2023srformer}, MARCONet~\citep{li2023learning}, MARCONet++~\citep{li2025enhanced}, DiffTSR~\citep{zhang2024diffusion} and our method.}
  \label{fig:realce_ours_x2_1}
\end{figure}

\begin{figure}[H]
  \begin{center}
    \includegraphics[width=\linewidth]{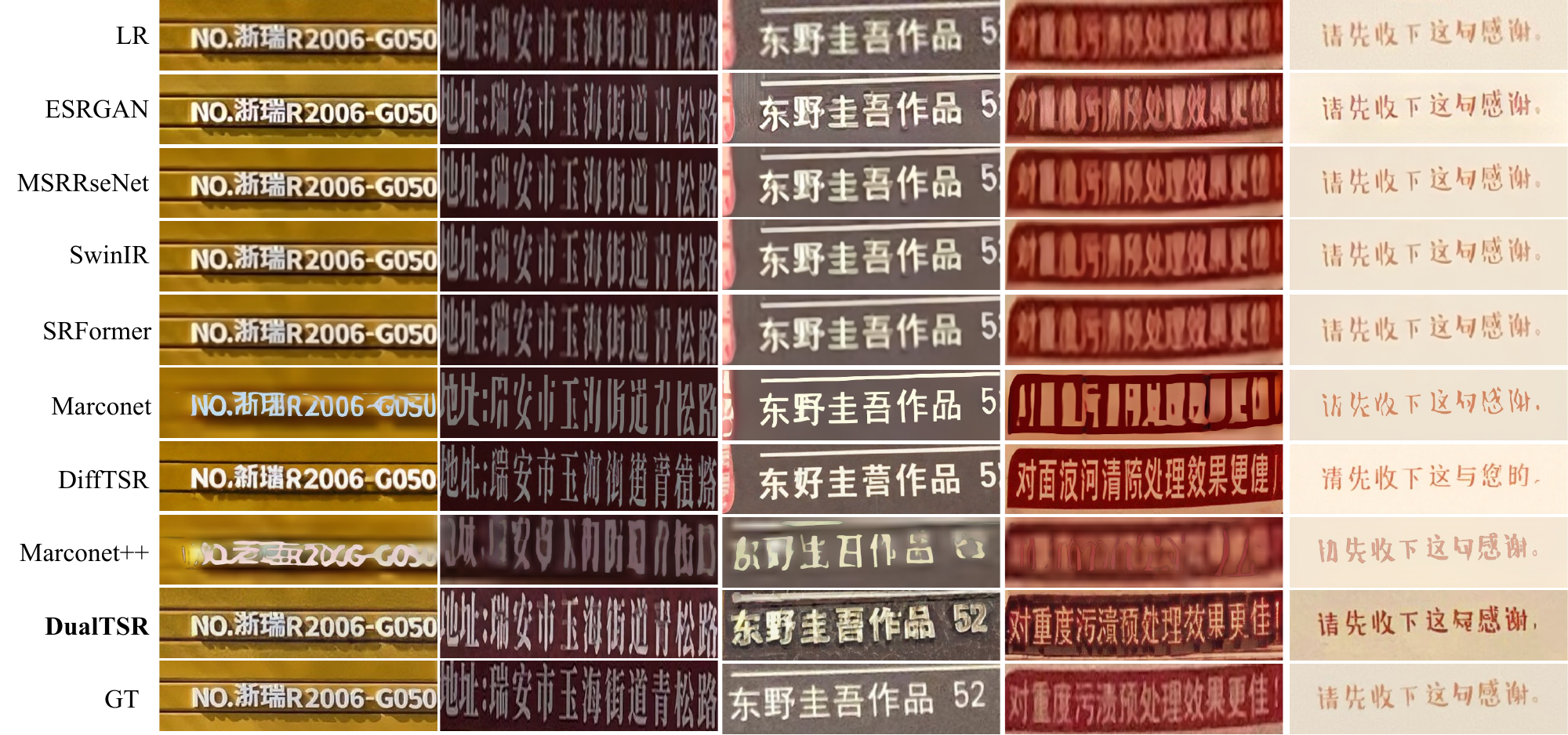}
  \end{center}
  \caption{\textbf{Visual comparison for the real-world dataset RealCE with different methods on $\times2$ scale.}
  The comparison methods include ESRGAN~\citep{wang2018ESRGAN}, MSRResNet~\citep{9022144}, SwinIR~\citep{liang2021swinir}, SRFormer~\citep{zhou2023srformer}, MARCONet~\citep{li2023learning}, MARCONet++~\citep{li2025enhanced}, DiffTSR~\citep{zhang2024diffusion} and our method.}
  \label{fig:realce_ours_x2_2}
\end{figure}

\begin{figure}[H]
  \begin{center}
    \includegraphics[width=\linewidth]{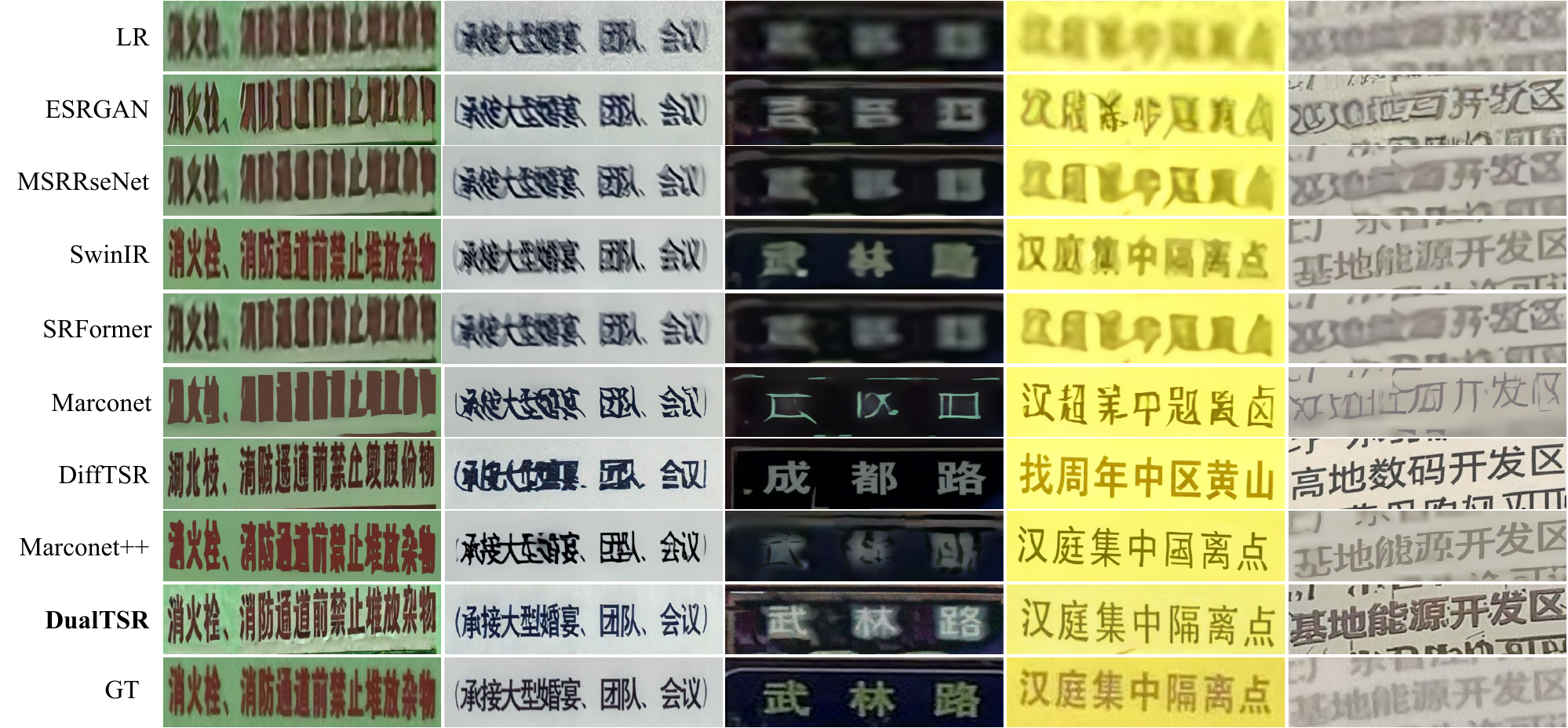}
  \end{center}
  \caption{\textbf{Visual comparison for the real-world dataset RealCE with different methods on $\times4$ scale.}
  The comparison methods include ESRGAN~\citep{wang2018ESRGAN}, MSRResNet~\citep{9022144}, SwinIR~\citep{liang2021swinir}, SRFormer~\citep{zhou2023srformer}, MARCONet~\citep{li2023learning}, MARCONet++~\citep{li2025enhanced}, DiffTSR~\citep{zhang2024diffusion} and our method.}
  \label{fig:realce_ours_x4_2}
\end{figure}

\begin{figure}[H]
  \begin{center}
    \includegraphics[width=\linewidth]{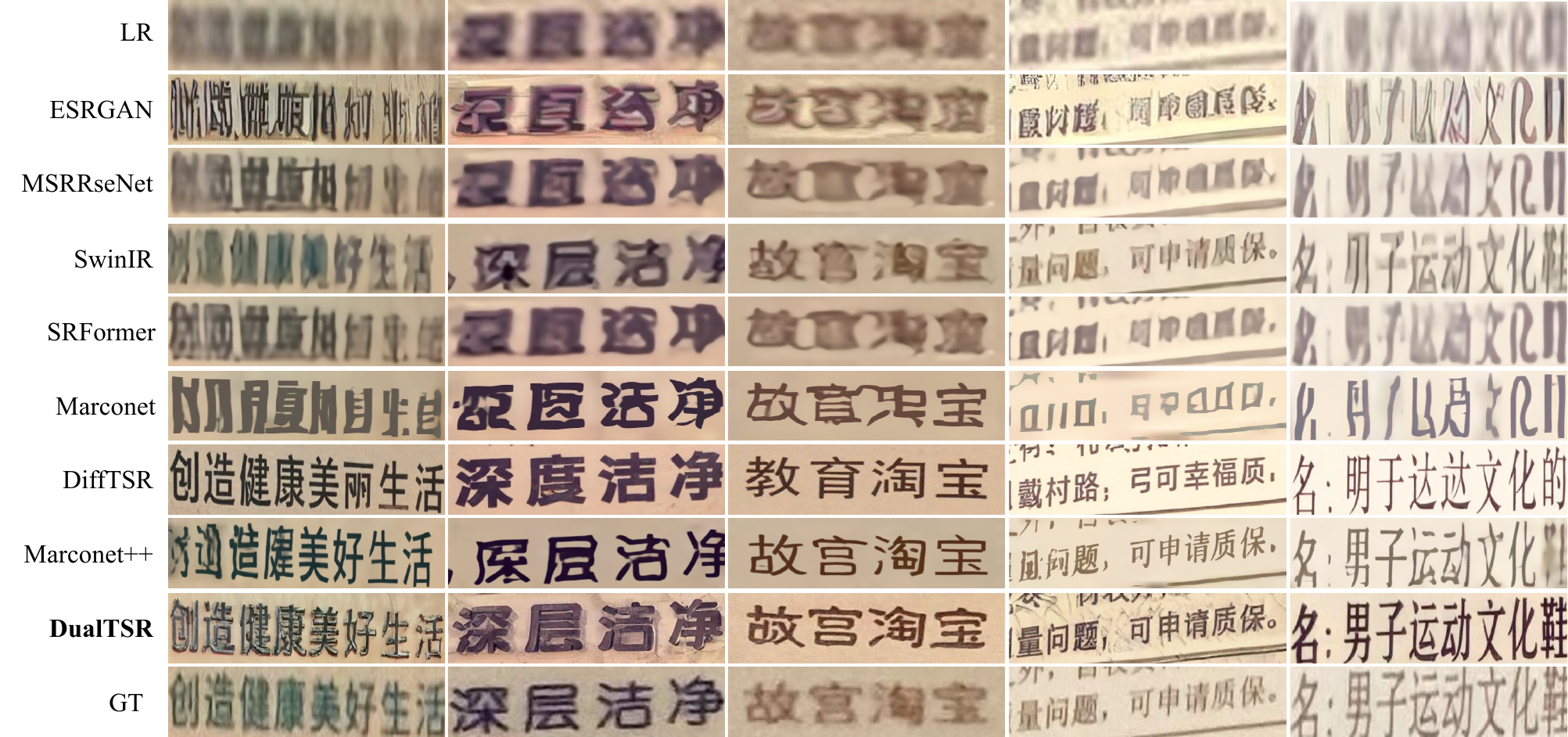}
  \end{center}
  \caption{\textbf{Visual comparison for the real-world dataset RealCE with different methods on $\times4$ scale.}
  The comparison methods include ESRGAN~\citep{wang2018ESRGAN}, MSRResNet~\citep{9022144}, SwinIR~\citep{liang2021swinir}, SRFormer~\citep{zhou2023srformer}, MARCONet~\citep{li2023learning}, MARCONet++~\citep{li2025enhanced}, DiffTSR~\citep{zhang2024diffusion} and our method.}
  \label{fig:realce_ours_x4_3}
\end{figure}

For the real-world RealCE benchmark, we provide additional $\times2$ results in Figures~\ref{fig:realce_ours_x2_1} and~\ref{fig:realce_ours_x2_2}, and additional $\times4$ results in Figures~\ref{fig:realce_ours_x4_2} and~\ref{fig:realce_ours_x4_3}.


\section{Training and Sampling Pseudocode}
\label{app:pseudo_code}
Here, we provide pseudocode for the DualTSR training process in Algorithm~\ref{alg:training} and the inference process in Algorithm~\ref{alg:inference}.

\input{sec/dual_tsr_alg}

%% file: sec/dual_tsr_alg.tex
\renewcommand{\lstlistingname}{Algorithm}

\begin{figure}[t]
\begin{lstlisting}[label={alg:training}, caption={PyTorch-style pseudocode for the proposed DualTSR training strategy. We utilize an EMA teacher model to provide guidance targets for the flow matching objective.}]
def train_step(model, model_ema, optimizer, batch, w=1.0, psi=0.1):
    x_hr, x_lr, text = batch  # Unpack data batch
    
    # 1. Image-Only Guided Loss (Flow Matching + Model Guidance)
    t = torch.rand(x_hr.shape[0])
    x_1 = torch.randn_like(x_hr)           # Sample Noise
    x_t = (1 - t) * x_hr + t * x_1         # Linear Interpolation
    
    # Calculate Guidance Target using EMA Teacher
    with torch.no_grad():
        out_cond = model_ema(x_t, t, cond={'img': x_lr, 'txt': text})
        out_unc  = model_ema(x_t, t, cond={'img': x_lr, 'txt': None})
        
        # Target = Base_Velocity + w * (Cond_Pred - Uncond_Pred)
        u_target = (x_1 - x_hr) + w * (out_cond - out_unc)

    # Student Prediction (with CFG dropout)
    cond = {'img': x_lr, 'txt': text} if rand() > psi else {'img': x_lr, 'txt': None}
    u_pred = model(x_t, t, cond)
    loss_img = F.mse_loss(u_pred, u_target)

    # 2. Text-Only Loss (Masked Generative Training)
    t_txt = sample_timesteps()
    text_masked = mask_tokens(text, ratio=1-t_txt)
    
    logits_txt = model(text_masked, t_txt, cond={'img': x_hr}, mode='text')
    loss_txt = F.cross_entropy(logits_txt, text) / t_txt

    # 3. Joint Guided Loss (Simultaneous Generation)
    t_j = torch.rand(x_hr.shape[0])
    x_t_j = (1 - t_j) * x_hr + t_j * x_1
    txt_t_j = mask_tokens(text, ratio=1-t_j)

    with torch.no_grad():
        # Joint guidance targets from EMA
        c_j = {'txt_in': txt_t_j, 'lr': x_lr}
        u_j_c = model_ema(x_t_j, t_j, c_j)
        u_j_u = model_ema(x_t_j, t_j, {'txt_in': None, 'lr': x_lr})
        u_target_j = (x_1 - x_hr) + w * (u_j_c - u_j_u)

    c_stud = c_j if rand() > psi else {'txt_in': None, 'lr': x_lr}
    u_pred_j, logits_j = model(x_t_j, t_j, c_stud, mode='joint')
    loss_joint = F.mse_loss(u_pred_j, u_target_j) + F.cross_entropy(logits_j, text)

    # Optimization Step
    loss = loss_img + loss_txt.mean() + loss_joint
    loss.backward()
    optimizer.step()
    update_ema(model, model_ema)
\end{lstlisting}
\label{alg:pytorch_code}
\end{figure}

\begin{figure}[t]
\begin{lstlisting}[ label={alg:inference}, caption={PyTorch-style inference algorithm. The image is generated via ODE flow integration, while the text is generated via iterative unmasking (discrete diffusion) synchronized with the image generation steps.}]
def joint_inference(model, x_lr, steps=50, seq_len=25):
    # 1. Initialization
    bs = x_lr.shape[0]
    x = torch.randn_like(x_lr)                     # Image Latent (t=1)
    text = torch.full((bs, seq_len), MASK_TOKEN)   # Text Latent (All Masked)

    # Time schedule from 1.0 down to 0.0
    timesteps = torch.linspace(1.0, 0.0, steps + 1)
    dt = 1.0 / steps

    # 2. Generation Loop
    for k in range(steps):
        t = timesteps[k]
        s = timesteps[k+1]  # Next timestep

        # Predict Velocity and Text Logits
        u_pred, logits_txt = model(x, t, text, cond=x_lr)

        # --- A. Image Update (Euler ODE Step) ---
        # Reverse the flow: subtract velocity to go from Noise -> Data
        x = x - dt * u_pred

        # --- B. Text Update (Reverse Discrete Diffusion) ---
        # Determine probability to unmask at this step
        alpha_t, alpha_s = 1 - t, 1 - s
        prob_unmask = (alpha_s - alpha_t) / (1.0 - alpha_t + 1e-8)

        # Sample candidate tokens from model prediction
        probs = F.softmax(logits_txt, dim=-1)
        candidates = torch.multinomial(probs.view(-1, probs.size(-1)), 1).view(bs, seq_len)

        # Update Logic:
        # If token is [MASK] AND random < prob_unmask -> Update with candidate
        # Else -> Keep current state (either already generated or still [MASK])
        is_masked = (text == MASK_TOKEN)
        should_sample = torch.rand_like(text.float()) < prob_unmask
        
        update_mask = is_masked & should_sample
        text[update_mask] = candidates[update_mask]

    return x, text
\end{lstlisting}
\label{alg:inference_code}
\end{figure}